\documentclass[Afour,sageh,times]{sagej}

\usepackage{moreverb,xurl}
\usepackage{lmodern}

\usepackage[colorlinks,bookmarksopen,bookmarksnumbered,citecolor=red,urlcolor=red]{hyperref}

\newcommand\BibTeX{{\rmfamily B\kern-.05em \textsc{i\kern-.025em b}\kern-.08em
T\kern-.1667em\lower.7ex\hbox{E}\kern-.125emX}}

\def\volumeyear{\the\year{}}
\def\journalname{Preprint}

\usepackage[justification=centering]{caption}

\begin{document}

\runninghead{Selecting Language Models for Social Science}

\title{Selecting Language Models for Social Science:\\Start Small, Start Open, and Validate}
\author{Dustin S. Stoltz\affilnum{1}, Marshall A. Taylor\affilnum{2}, and Sanuj Kumar\affilnum{3}}

\affiliation{
\affilnum{1}Department of Sociology and Anthropology, Lehigh University\\
\affilnum{2}Department of Sociology, New Mexico State University\\
\affilnum{3}Department of Computer Science, New Mexico State University\\
}

\corrauth{
Dustin S. Stoltz, 
Department of Sociology and Anthropology, Lehigh University}
\email{dss219@lehigh.edu}

\begin{abstract}
Currently, there are thousands of large pretrained language models (LLMs) available to social scientists. How do we select among them? Using validity, reliability, reproducibility, and replicability as guides, we explore the significance of: (1) model openness, (2) model footprint, (3) training data, and (4) model architectures and fine-tuning. While \emph{ex-ante} tests of validity (i.e., benchmarks) are often privileged in these discussions, we argue that social scientists cannot altogether avoid validating computational measures (\emph{ex-post}). Replicability, in particular, is a more pressing guide for selecting language models. Being able to reliably replicate a particular finding that entails the use of a language model necessitates reliably reproducing a task. To this end, we propose starting with smaller, open models, and constructing delimited benchmarks to demonstrate the validity of the entire computational pipeline.
\end{abstract}


\maketitle

A language model predicts words given the presence of other words. OpenAI's tools are the prototypical encounter with large pretrained language models (LLMs) and generative artificial intelligence (GenAI). The broader ecosystem, however, is wide and varied, with thousands of pretrained models available.\footnote{Roughly 1 to 2 thousand models are uploaded to the repository Hugging Face every day (\url{https://www.interconnects.ai/p/2025-open-models-year-in-review})} Google's Gemma, for instance, can be downloaded and used locally on any computing platform without sending information to Google's servers. Furthermore, some non-profit and researcher coalitions like EleutherAI or the Allen Institute are also contributing models and training materials, providing alternatives that can potentially address some of the ethical and technical shortcomings of popular LLMs like OpenAI's GPT models.

The use of LLMs in published sociological research is currently sparse.\footnote{There are a handful of proposed uses of LLMs in sociological research \citep{Davidson2025-kf}. However, a recent survey shows that, to the extent that researchers in sociology are using LLMs (or other generative ``AI" tools), they are using them primarily as grammar and spellcheckers, paraphrasers, or to assist with writing and debugging programming code---that is, as procedural aids \citep[]{Alvero2025-kl}} We suspect much will change as norms develop around the role, if any, of LLMs in research workflows. This paper is a contribution to this discussion. While some previous research provides guides for applying LLMs to specific research tasks, we take a task-agnostic approach. That being said, the following applies to the class of tasks that analysts would like to \emph{iterate} and where such iteration directly relates to scientific explanation.\footnote{For example, using an LLM to generate article titles or synonyms for a word in an abstract would not fall into this category.} So, assuming we have a research task for which an LLM is an appropriate tool, how should we choose among them?\footnote{Furthermore, we do not consider the use of various front-end ``graphical-user interfaces" (GUIs). Our focus is squarely on selecting the model. However, many GUI tools are being developed that can use any model on the backend, such as the open-source GPT4ALL (\url{www.nomic.ai/gpt4all}) or Librechat (\url{https://www.librechat.ai/}).}

How do we select among language models? Using reproducibility, reliability, validity, and replicability as our guides, we explore four categories of consideration when selecting language models: (1) model openness, (2) model footprint, (3) training data, and (4) model architecture and fine-tuning. The first comprises considerations about the transparency of training procedures and model architectures, the extent to which model weights are available to download and inspect, and whether users have control over how a model is trained, operated, and where information is stored. The second entails considerations about the number of parameters, whether a model is a distillation, the level of quantization/precision of weights, and the context length of the model. The third relates to who is involved in training the model, how much and what kind of data a model is trained with, and the ethics of the training materials used. The last considers different model architectures, whether they have been trained on datasets to fine-tune them for a particular task (i.e., instruction-following, preference-alignment, reasoning), and retrieval-augmented language models.

\section{BACKGROUND}

\subsection{Anatomy of Language Models}

Language models have been around for decades and generally refer to algorithms that predict a \emph{probable next token} (characters, words, phrases, or sentences) given previous tokens (or characters, words, phrases, or sentences). There are two approaches to language modeling: (1) rule-following or symbolic models based on grammar and decision-trees; and (2) probabilistic or statistical modeling based on the observation of actually occurring instances of language use. The models we'll discuss build on this probabilistic tradition, specifically using neural networks \citep{Bengio2003-ps}. 

Earlier language models faced two hurdles. First, most sequences of words are quite rare when we look at sequences of even three or four words. This necessitates the use of very, very large corpora to get accurate probabilities. The infrastructure for the collection and access to large amounts of text has only been enabled by the Internet and the many people who have written blog posts, made social media comments, uploaded digitized books, or updated Wikipedia articles. Second, words are not only constrained by their immediate context. Text has ``long-range" dependencies, which might make a word more or less probable. There were several incremental improvements on this front that led to the long short-term memory (LSTM) architecture \citep{Hochreiter1997-qy} and the attention mechanism \citep{Bahdanau2014-lb, Lin2017-tj}, both culminating in the ``Transformer" which is almost universally used in current large language models (see Gu and Dao (\citeyear{Gu2023-ml}) and Beck et al. (\citeyear{Beck2024-vb})).\footnote{The Transformer consists of ``blocks" that each perform two steps: (1) multi‑head self‑attention and a (2) position‑wise feed‑forward network. The attention mechanism computes weighted combinations of all words in the sequence so that the model can focus on the most relevant parts of the input. Positional encoding is added because the network itself does not understand word order. Transformers process entire sequences in parallel and capture long‑range dependencies.}

Here, we focus on selecting \emph{pretrained} Transformer-based language models. For our purposes, each pretrained language model has two main components: (1) the tokenizer, input embeddings, and position embeddings, and (2) the pretrain weights. The latter consists of a series of blocks, and within those blocks are layers of matrices---the model is, therefore, often referred to as a metaphoric ``stack." There are also rules, predefined by the model architecture, that define how information is passed between layers and blocks. The weights, by contrast, are ``learned" during the training process. Most LLMs \citep[]{Raffel2020-ox} are first trained on very large ``generic" corpora (like the Common Crawl or The Pile), typically involving a task like next word, masked word, or next sentence prediction. The result is sometimes referred to as a ``foundation model" \citep{Bommasani2021-vq}.\footnote{This is sometimes mixed up with ``frontier" model, which is more-so a marketing term used to refer to whatever LLM is the newest, meant to evoke continual progress.} These foundation LLMs are then typically trained on more specific datasets (i.e., post-training or fine-tuning) with tasks like paraphrasing, question-answering, sentiment analysis, or translation. Finally, to make the model ``generative" in the narrow sense, an explicit ``language modeling" block can be added after the weights that converts the output into exactly one token.\footnote{This is usually a linear layer and then the softmax function selects the most probable token.}

\subsection{The Organizational Landscape of LLMs}

Since the launch of OpenAI's GPT-2, money and resources devoted to AI have expanded considerably. Since 2020, global corporate investment in AI broadly jumped US\$ 100s of billions \citep[]{Maslej2025-ho}. In 2024, there were over 2,000 newly funded companies, 10\% of which focused on ``generative AI" \citep[250]{Maslej2025-ho}. 

OpenAI is probably the most well-known organization building LLMs today. It has a private organizational structure that started with a non-profit at its core, and several for-profit subsidiaries. These for-profits allow the organization to obtain investments. However, in late 2025, OpenAI converted to a for-profit public benefit corporation, but with the non-profit OpenAI Foundation holding 26\% stake in the company.

Microsoft is the largest investor in OpenAI, a major shareholder in the forprofit, and initially the sole provider of cloud computing resources through its Azure service. This means that when interacting with ChatGPT through a web browser, the actual inference computing is conducted (primarily) on Microsoft's hardware. OpenAI uses a ``freemium" strategy, which offers limited services for free, with the option to pay for different levels of access---and OpenAI may be planning to push advertisements \citep[e.g.,][]{Cuthbertson2025-mt}.\footnote{Importantly, pricing does not necessarily reflect total compute costs or specific profit targets. OpenAI has lost \$US billions per year \citep{Isaac2024-fr}. These prices are also subsidized by Microsoft and several venture capital firms. These investors include Sequoia and Andreessen Horowitz (\url{https://pitchbook.com/news/articles/top-generative-ai-vc-investors-list}). In addition to cheaper hardware and more efficient ways to operate LLMs, the investors likely hope that OpenAI will ``lock in" users into the product interface. This may happen because, for example, users have a collection of prompts that work well with ChatGPT but not other LLMs \citep{Hosanagar2024-nl}. As OpenAI's models are proprietary, users can only access them through ChatGPT or OpenAI's API. Thus, once users are ``locked in," they could eventually be charged more for the same level of access. Additionally, other companies may build application ``frontends" that work on top of OpenAI's model.}

\begin{table*}[]
\centering
\begin{tabular}{p{4cm}|c|c|c|c}
\toprule[1.5pt]
               &     Model          &  Open-Weight &     API       &   GUI        \\
Organization   &     Family         &  Available   &   Access     & Access        \\ \hline
OpenAI         &    GPT             &  \checkmark  & \checkmark  &  \checkmark   \\
Anthropic      &    Claude          &              & \checkmark &  \checkmark   \\
Meta           &    Llama           &  \checkmark  & \checkmark &  \checkmark   \\  
Google         &    Gemma/Gemini    &  \checkmark  & \checkmark &   \checkmark    \\    
Mistral        &    Mistral/Mixtral &  \checkmark  & \checkmark  &  \checkmark   \\   
Microsoft      &    Phi             &  \checkmark  &              &               \\   
Alibaba        &    Qwen            &  \checkmark  & \checkmark   &  \checkmark   \\   
Cohere         &    Command         &              & \checkmark   &               \\   
AI21           &    Jamba           &  \checkmark  & \checkmark   &              \\   
DeepSeek       &    DeepSeek          &  \checkmark  & \checkmark   &  \checkmark  \\   
Apple          &    OpenELM           &  \checkmark  &              &              \\   
IBM            &    Granite           &  \checkmark  &              &              \\   
EleutherAI     &    Pythia            &  \checkmark  &              &              \\  
AllenAI (Ai2)  &    OLMo              &  \checkmark  &              & \checkmark   \\  
01-ai          &    Yi                &  \checkmark  &              &              \\  
Databricks     &    DBRX              &  \checkmark  &              &              \\  
MoonshotAI     &    Kimi              &  \checkmark  &              &              \\  
Hugging Face   &    Bloom/Smol        &  \checkmark  &              &              \\  
Nvidia         &    Nemotron          &  \checkmark  & \checkmark  & \checkmark  \\  
Falcon Foundation &    Falcon         &  \checkmark  & \checkmark   & \checkmark   \\  
Swiss Nat'l AI Initiative & Apertus   &  \checkmark    &              &    \\  
Zyphra & Zamba/BlackMamba &  \checkmark    &              &    \\  
LLM360 & K2/Crystal &  \checkmark    &              &    \\  
\bottomrule[1.5pt]
\end{tabular}
\begin{center}
\caption{Selection of Model Families}\label{tab:terms} 
\end{center}
\end{table*}

Meta (formerly Facebook) is also a major player in the GenAI space, releasing one of the most well-known \emph{open-weight} models called Llama, which we can download, and with some setup, use this for inference directly on the computing hardware provided by a desktop, laptop, or even a smartphone \citep{Zhang2025-yy}. We could also use it on a university high-performance computing cluster (HPC), or purchase access to cloud computing servers, namely Microsoft's Azure, Amazon's AWS, and Google's Cloud.\footnote{Market research group Synergy \citep*{Synergy-Research-Group2025-ob} estimates that these three companies account for over half of the cloud computing market share, respectively, at 30\%, 21\%, and 12\%. Other providers include Alibaba, Oracle, IBM, and Tencent. Many of these companies offer so-called ``serverless" options, where we wouldn't pay for server space or computing power when we're not using it and offer a ``pay-as-you-go" billing model.} If an organization offers either API or GUI access, then inference will be processed on the organization's hardware (or the servers they rent). A benefit of open-weight models is that, as researchers, we can exercise greater control over data storage.\footnote{Data retention practices for LLM companies is an evolving legal landscape (e.g., \url{https://openai.com/index/response-to-nyt-data-demands/}. For regular users of ChatGPT, prompts may potentially be used by OpenAI to train later models or fine-tune existing models. But, they do allow individual's to opt-out (\url{https://openai.com/policies/how-your-data-is-used-to-improve-model-performance/}). For Enterprise services (such as those offered to educational institutions), the organizations must opt-in (\url{https://openai.com/enterprise-privacy/}, and this includes users of the API. OpenAI still retains deleted prompts for 30 days. And, of course, all of this assumes a level of trust in the organization.}

Open weight models are typically downloadable from the websites of the organizations that train them (e.g., Meta, Google, Mistral, and so on). However, there are also repositories that host open-weight models, namely Hugging Face, Ollama, Docker Hub, GitLab, Kaggle, ModelScope, and GitHub---all owned by commerical enterprises.\footnote{The first five are all owned by companies of the same name. ModelScope was launched by the Alibaba Group in 2022; GitHub was acquired by Microsoft in 2018; and Kaggle was acquired by Google in 2018.} Two possible non-profit alternatives include Codeberg, which is a repository service (established in 2019) and Zenodo, a general-purpose repository operated by CERN (European Organization for Nuclear Research) that is used primarily by academics to post datasets and replication code. A selection of organizations and their language model families is in Table \ref{tab:terms}.

Finally, we should consider the hardware, specifically the semiconductor components. Whether we use a remote server, an HPC, or our own laptop, we rely on processing chips---either the Central Processing Unit (CPU) or the Graphical Processing Unit (GPU). Nvidia has over 90\% of the GPU market share \citep{Morales2025-rr} (and 35\% of the semiconductor market more broadly) due to the current efficiencies of their GPUs for training and inference. Nvidia relies heavily on TSMC's chips, which has roughly 70\% of the chip market share \citep{James2025-ge}. In turn, ASML is the sole provider of a key technology involved in producing these chips \citep{Elstrom2025-rr}. While these market shares or organizations may change in the coming years, there is likely to remain a considerable amount of market concentration in the hardware that supports GenAI.

\section{BENCHMARKS}

Comparing an algorithm or machine learning model's performance on established ``benchmarks" has been a standard practice for decades---in part to evade the boom and bust of attention bubbles. Spurred by work in automated translation and speech recognition research, John Pierce argued that a problem with the research is that it was ``attractive to money" \citep[]{Pierce1969-hl}:

\begin{quote}
To sell suckers, one uses deceit and offers glamour. It is clear that glamour and any deceit in the field of speech recognition blind the takers of funds as much as they blind the givers of funds.
\end{quote}

This incentivized the ``mad inventor" over the incremental ``progress" of the scientist. To rise to Pierce's challenge, researchers in the field turned to quantitative comparisons in formal competitions \citep{Donoho2017-kh}. In theory, having a collection of public benchmarks seems like an obvious boon. However, we should remember the Goodhart-Campbell Law: ``When a measure becomes a target, it ceases to be a good measure" \citep[]{Goodhart1984-jf, Campbell1979-ne}. In the case of evaluating machine learning models, GenAI in particular, we have reason to be even more apprehensive, since the evaluators stand to gain from the evaluations: ``[O]veremphasizing metrics leads to manipulation, gaming, a myopic focus on short-term goals, and other unexpected negative consequences" \citep{Thomas2020-wk}

There are now numerous examples of companies and researchers \emph{overpromising} on the capabilities of machine learning models---LLMs and other generative models included. For example, AI tools might speed up processes in, say, radiology, but several studies show that the time-savings are either minimal (a few seconds) or non-existent \citep[]{Kim2022-ll, Shin2023-np, Muller2022-ti}. Or, consider a highly publicized paper \citep[]{Merchant2023-ku}---published in \emph{Nature} and authored by a team from Google's DeepMind---that claimed to have discovered 2.2 million materials and synthesized 43 ``novel materials," which ``represents an order-of-magnitude expansion in stable materials known to humanity." In a review \cite[3490]{Cheetham2024-qd} of these ``new materials," however, researchers found ``scant evidence" that anything new or useful was actually discovered \cite[see also ][]{Leeman2024-kq, Magesh2025-qx}. The common pattern of inaccurately reporting model performance should leave us justifiably skeptical.

The source of these outsized claims can be traced to numerous questionable practices \citep{Leech2024-ph}. We will focus on three key issues related to current LLM benchmarking practices: (1) competing baselines, (2) benchmark leakage, and (3) construct validity. Before discussing each, let's breakdown benchmarks.

\subsection{Anatomy of a Benchmark}

Benchmarks are datasets (or sets of datasets) that include inputs and desired outputs corresponding to a specific task, and some method to assess performance. While there are thousands of benchmarks used to test LLMs, there are a few well-known (if superseded) examples such as GLUE (Generalized Language Understanding Evaluation, \cite{Wang2018-nx}) and MMLU (Massive Multitask Language Understanding) that other benchmarks tend to be modelled on. In contrast to, for example, how a part-of-speech benchmark might measure a part-of-speech tagger's performance, these benchmarks are foundational in that they attempt to measure how well a language model performed on supposedly ``general" or ``out of domain" tasks.

To accomplish this, GLUE incorporates nine previously created benchmarks. For example, the SST-2 task (the Stanford Sentiment Treebank Binary task) involves classifying whether a movie review (the input) is positive or negative (a binary output), and assess performance with \emph{accuracy} (the correct predictions divided by the total). Or consider the Stanford Question Answering Dataset, where the model determines (a yes/no output) whether a paragraph drawn from Wikipedia contains the answer to a corresponding question, and again reports the accuracy. As a final example from GLUE, the STS-B (Semantic Textual Similarity Benchmark) contains pairs of sentences and human ratings of similarity from 1 to 5. Performance on this task is the correlation (Pearson or Spearman) between those similarity ratings and the models' ratings. The final performance is the average across all of these nine sub-benchmarks. Some benchmark examples are provided in Table \ref{tab:bench}.

Verification of a ``correct" response can be more complicated, as in the case of ``instruction following" benchmarks. For instance, IFEval \citep{Zhou2023-sz} prompts models to produce an output with specific requirements such as using a specific number of uppercase characters, a specific character length, or ending the response with a specified phrase. To verify the prompt ``Write a riddle for teens about hospitals but don't use any punctuation," we would use a basic string matching search for punctuation. Or, for ``Write a dad joke about funerals in Azeri, no other language is allowed," a language detection algorithm would be used. To account for near matches, often verification will involve fuzzy matching using ``edit distance" \citep{Navarro2001-oh}, text similarity measures, such as BLEU \citep{Papineni2002-lc}, ROUGE \citep{Lin2004-um}, or structural similarity measures like TSED \citep{Song2024-si}.

\begin{table}[]
\centering
\begin{tabular}{c|p{4cm}|p{2cm}}
\toprule[1.5pt]
Dataset & Example                                           &  Task             \\ \hline
QNLI    & Q: What is the Rankine cycle sometimes called?

          A: The Rankine cycle is sometimes referred to 
             as practical Carnot cycle.                     &  Question Answering  \\ \hline
GSM8K   & Q: Jimmy decides to run 3 sprints 3 times a week.  
             He runs 60 yards each sprint. How many 
             meters does he run a week?
             
          A: He sprints 3*3=<<3*3=9>>9 times
             $\backslash$nSo he runs 9*60=<<9*60=540>>540 
             yards$\backslash$n\#\#\# 540               &  Math Reasoning   \\ \hline
IFEval  & Q: Write a quiz about gigabytes that includes the 
             word pangolin at least 3 times.              &  Instruction Following \\

\bottomrule[1.5pt]
\end{tabular}
\caption{Benchmark Examples}
\label{tab:bench} 
\end{table}

\subsection{Competing Baselines} 

When demonstrating that a model is superior in some respect, we often see it compared directly to other models or techniques. These other models/techniques provide a \emph{baseline}; if one's model scores higher, one can claim some improvement, however small. There are at least three potential pitfalls. First, aside from cases where there is an obvious leading champ (i.e., SOTA or State-of-the-Art), one typically cherry-picks the set of competitors. Second, one usually re-scores the competing models/techniques at the same time they run their own, preferred model. Third, the evaluators can also cherry-pick which benchmarks they choose to report.

A team \citep{llmrl2025incorrect} attempted to replicate the claims from seven papers arguing that ``reinforcement-learning" (or RL, a procedure discussed later) improved the ``reasoning abilities" of LLMs. Each paper showed purported performance ``gains" on benchmarks before and after RL. The replicators, however, failed to see improvements because ``baseline numbers of the pre-RL models are significantly underreported," compared to previously reported scores. Another team \citep{McGreivy2024-om} found that, out of 76 papers that claimed to solve ``partial differential equations" faster with machine learning than with numerical methods, 60 underreported the baselines of the numerical methods. 

With LLMs, this could result from variation in hyperparameters such as the exact wording of the prompt, context-length cut-offs, or issues with parsing the output leading to false negatives or false positives. It may also result from slight variations in the setup, such as the order in which tasks are completed \citep{Gupta2024-qw, Xu2024-gq}.

Finally, developers could cherry-pick benchmarks more favorable to their preferred model. This is a problem for more recent concerns about ``factuality" and especially for tasks that are difficult to unambiguously define, such as ``responsible" model outputs \citep{Maslej2024-zk, Maslej2025-ho}.

\subsection{Benchmark Leakage}

Leakage refers to any feature of the evaluation that ``allows a model to `cheat' by using information that it should not have access to" \citep[4]{Sculley2025-pd}. This is not a problem solely with LLMs. For instance, one team of computer scientists \citep{Kapoor2023-gv} found hundreds of machine learning papers in numerous fields encountered benchmark leakage of some kind, resulting in overoptimistic results \citep[see also][]{Xu2024-gq, Sasse2025-uz, Deng2023-ve, Zhou2025-hf}. 

Training LLMs requires a large amount of data. These data are collected and vetted using automated processes and are often so large that trainers cannot be entirely sure what is included. Often this may include text that is very similar to the benchmarks, or text from which benchmarks are constructed, such as Reddit or Wikipedia \citep{Haimes2024-ph}. In some cases, the actual benchmarks may be included in training materials---a form of leakage called \emph{contamination} \citep[]{Magar2022-vw}. Even in so-called ``chatbot arenas" some providers---OpenAI, Meta, and Google---are able to access more data from the arenas and test multiple variants in ways that game the leaderboard rankings (\cite{Singh2025-sh}; see also \cite{Balloccu2024-be}). 

This is only part of the problem, however. There is also the question of \emph{memorization}. Language models may not be ``learning" general information that is used to solve a task, but actually memorizing the training text. For example, in the math reasoning benchmark GSM8K, one question includes the text: ``Jerica is twice Louis' age." If we prompt an LLM with just ``Jerica is" and the model completes the n-gram ``twice Louis' age." exactly, it is very likely the model was trained on the benchmark and ``memorized" the question. Xu et al (\citeyear{Xu2024-gq}) find, for instance, that Qwen-1\_8B accurately generated all 5-grams in 223 questions from the GSM8K \citep{Cobbe2021-js}. In this case, the benchmarks are not measuring what we think they are.

As possible solutions, some have proposed avoiding posting benchmarks as publicly-available plain text documents \citep{Jacovi2023-nb}; including a ``canary string" \citep{Srivastava2022-fp} or a unique string of characters with benchmarks that can be used to filter out web pages with benchmarks or be used to detect contamination at the time of inference; or creating multiple variations on acceptable answers, only one of which is selected for the benchmark \citep{Ishida2025-hl}.

\subsection{Construct Validity}

Validity refers to how well a given metric measures what we want it to measure. Whereas previous natural language processing models were trained and tested on specific tasks, such as part-of-speech tagging or coreference resolution, LLMs are trained and tested on the ability to perform multiple-tasks without altering the model much, if at all. 

As discussed, current benchmarks are typically combinations of datasets that attempt to test how the model performs on specific tasks. They are often quite large and, furthermore, researchers often borrow previous benchmarks, many of which are outside the researchers' own expertise. Furthermore, as dataset construction is often poorly documented, this combination leads to a situation in which current benchmarks often contain errors, either because the researchers do not have the resources or the expertise to double-check the datasets.

A second issue, however, is that the notion of a ``general" ability may be incoherent. All datasets are ``limited in ways that necessitate a different framing from that of any claim to general knowledge or general-purpose capabilities" \citep{Raji2021-vs}. For instance, the MMLU includes a task on ``moral" decision-making in which two scenarios are provided, and then the model is tasked with stating whether each scenario is wrong or right (morally speaking). But, moral according to whom? This task was borrowed from the ETHICS benchmark \citep{Hendrycks2020-gz}, and these researchers copied text from Reddit and also used the crowdsourcing platform Amazon Mechanical Turk. Crowdsource workers were tasked both with writing some scenarios and also judging scenarios. The researchers use agreement between crowdsource workers as a measure of ``ambiguity," and included examples where there was 95\% agreement (as clear cut examples) and 50\%$\pm$10\% (as ambiguous). Although the paper is unclear, this appears to be the initial agreement between 5 to 7 workers in the U.S., Canada, and Great Britain, and later agreement with 10 workers in India. While we can appreciate the steps the benchmark creators take to get at ``generality," the small number of annotators strains the plausibility that this gets at a ``general" moral faculty---if such a thing even exists.

Benchmarks are fundamentally limited, yet some researchers will nevertheless ``elevate them to the status of a target the entire field should be aiming for" \citep{Raji2021-vs}. Even if we see our particular social scientific task as a subset of some general capacity, it is likely that the benchmarks used to measure such a capacity involve contexts that are very different from our own. While the machine learning community is working to improve this state of affairs---e.g., adding error bars \citep{Miller2024-mj}, drawing explicitly on measurement theory \citep{Wallach2025-xh}, including a larger range of stakeholders in benchmark development \citep{Thomas2020-wk}, using continuously updated ``dynamic" benchmarks \citep{Jain2024-pc, Xu2024-gq}, or reporting disaggregated results \citep{Burnell2023-ju}---none of this will lead to a situation where social scientists can altogether avoid validating language models for our delimited social scientific task. In order to be confident that our computational pipeline is accomplishing what we think it is, we'll need to construct our own specialty benchmarks. We presume that these specialty benchmarks will generally take the form of hand-annotated subsets of the analysts' corpora---``validation sets" in supervised learning lingo.\footnote{The Oxford Internet Institute's Reasoning with Machine's Lab (OxRML) provides a handy tool to help determine the construct validity of benchmarks (\url{https://oxrml.com/measuring-what-matters/checklist.html?state=0}), based on a review of 445 benchmarks \citep{Bean2025-oi}.}

\section{MODEL OPENNESS}

\emph{Start open} is where we advise all social scientists to begin. But what makes a model ``open"? Is it just open weights (and what does that mean, anyway)? Where can we start if we start open?

\subsection{Open-Source or Open-Weight?}

When OpenAI announced GPT-2 in February 2019, they withheld much information about the model, including choosing not to release the full 1.5 billion parameter model \citep[]{Solaiman2019-if}. This was in contrast to previous models produced by OpenAI (e.g., GPT-1). Reporting from \emph{The Verge} explains: ``OpenAI is treading cautiously with the unveiling of GPT-2. Unlike most significant research milestones in AI, the [OpenAI] lab won't be sharing the dataset it used for training the algorithm or all of the code it runs on..." \citep{Vincent2019-jh}. Instead, OpenAI slowly released \emph{smaller} versions of the model, before finally releasing the full model. Despite OpenAI's initial reluctance, using the technical reports released by OpenAI alone, independent researchers were able to create OpenGPT2.

By the release of GPT-4 in early 2023, OpenAI's tune had changed considerably. The technical report introducing the model primarily discusses the model's performance on various benchmarks and little else. As the authors state: 

\begin{quote}
Given both the competitive landscape and the safety implications of large-scale models like GPT-4, this report contains no further details about the architecture (including model size), hardware, training compute, dataset construction, training method, or similar.
\end{quote}

When asked to clarify about the stance, the former chief scientist at OpenAI doubled-down \citep[]{Vincent2023-pb}:

\begin{quote}
We were wrong. Flat out, we were wrong. If you believe, as we do, that at some point, AI-AGI [Artificial General Intelligence] is going to be extremely, unbelievably potent, then it just does not make sense to open-source. It is a bad idea... I fully expect that in a few years it's going to be completely obvious to everyone that open-sourcing AI is just not wise.
\end{quote}

OpenAI changed it's tune, again, in 2025---to a degree. In August of that year, they released GPT-5 and GPT-OSS---the latter being a set of two new open-weight models. GPT-OSS (available in 117b and 21b total parameter flavors) is OpenAI's first open-weight release since GPT-2 and readily available in model repositories such as HuggingFace under the Apache 2.0 license. While the models themselves are open, the \emph{information} behind their architecture and training is not. OpenAI's ``model card"---a published statement meant to document the architecture, training process, benchmark performance, use cases, and other details about an AI/machine learning model---specifies only that GPT-OSS is built upon the GPT-2 and GPT-3 architectures and the training data consists of ``trillions of tokens, with a focus on STEM, coding, and general knowledge" \citep[]{openai2025gptoss120bgptoss20bmodel}.

This points us toward an important distinction in discussions around model openness: \emph{open weights} versus \emph{open source} \citep[]{widder2023open,bhandari2025forecasting}. The former refers to the open release of the model's data-learned numerical parameter space that maps model inputs to outputs and which can be fine-tuned for user-specific tasks; the latter refers to the full release of the model's entire production workflow---source code, training data (more on this later), weights, and more.

In contrast to OpenAI's stance regarding GPT-2, Meta released the Llama family of models in early 2023 \citep[]{Touvron2023-wg}. Although initially referred to as ``open source" models, this sparked a debate in the machine learning community. According to the Open Source Initiatives ``Open Source AI Definition,"\footnote{https://opensource.org/ai/open-source-ai-definition} a model must grant the following freedoms to be considered open source:

\begin{itemize}
  \item Use the system for any purpose and without having to ask for permission.
  \item Study how the system works and inspect its components.
  \item Modify the system for any purpose, including to change its output.
  \item Share the system for others to use with or without modifications, for any purpose.
\end{itemize}

Meta places some restrictions as to how their models can be used, which means it is not \emph{truly} open-source (according to the fourth criterion). Even then, Llama meeting the first three criteria relies on making an equivalency between ``model" and ``system"---e.g., that modifying the \emph{system} for any purpose is synonymous with modifying the \emph{model} for any purpose. This equivalency is a useful rhetorical device: under this definition of open-source, access to training data and training procedures is not assumed; rather, it is focused on the \emph{learned weights} of the pre-trained model. But is a model a \emph{system}? In the case of LLMs, not in any meaningful sense: it's not a stretch to argue that ``studying a model" (criterion 2) involves understanding the model's underlying training data or procedures, underscoring that being ``open-source" necessitates a systems-level---not a weight-level---view on a model. From a market vantage point, though, the ``open-source" label carries real consequences: under the new EU Artificial Intelligence Act\footnote{https://artificialintelligenceact.eu/}, models released with an open-source license may be exempt from new reporting and oversight regulations. This contributes to what is known as ``open-washing"---leverarging a discourse of model openness for the reputational and/or legal benefit but without the ``meat" that makes a model fully open-source \citep[p.~1776]{liesenfeld2024rethinking}. In attempt to hold developers accountable, the Stanford based Center for Research on Foundation Models continually evaluates foundational models along a 100 ``transparency indicator," with IBM being the most transparent \citep{Wan2025-bh}.

\subsection{Fully-Open Options}

Fully-open LLMs are much rarer than their closed- or open-weight counterparts. A popular family of fully-open LLMs is Pythia---16 models of various sizes released in 2023 \citep{biderman2023pythiasuiteanalyzinglarge}. The Pythia models are the product of EleutherAI---a non-profit organization with an eye toward being the open-source alternative to OpenAI's secrecy. Each model was trained using the exact same data in the same order, ``enabling researchers to explore causal interventions on the training process."\footnote{https://github.com/EleutherAI/pythia?tab=readme-ov-file} In addition to the model, the code used to train the models is publicly available as was\footnote{The Pile is no longer directly downloadable from EleutherAI due to copyright issues. But a newer version, The Common Pile, is now available, which we discuss in the ``Training Data" section of this paper.} the 825 GiB public corpus called \emph{The Pile}, also developed by EleutherAI \citep{gao2020pile800gbdatasetdiverse}. The suite also includes model checkpoints so researchers can examine the model outputs at difference stages in the training process. Furthermore, the results in the accompanying paper were independently verified. This effort created the blueprint for open LLMs. Other non-profit or government-led consortiums have since followed suit, including the Allen Institute for AI (OLMo models), LLM360 (K2 and Crystal models) and the Swiss National AI Initiative (Apertus) \citep{liu2025llm360k2building65b,groeneveld2024olmoacceleratingsciencelanguage,olmo20252olmo2furious, Apertus2025-ny}. The Falcon 180B, released by the Abu Dhabi government-funded Technology Innovation Institute, is one of the largest open-source models currently available \citep{almazrouei2023falconseriesopenlanguage}. Zyphra is one of the few for-profits releasing fully open models (Zamba, BlackMamba).

There is clearly a growing interest in developing open LLMs---a leaked memo from Google even claimed that open-source models are ``quietly eating our [Google's] lunch" \citep{Roth2023-wa}. We are also seeing the collective action necessary to see some of these developments through \citep{Phang2022-ke, Linaker2025-it, Tan2025-dd}. But, given the scale of financing needed to train a single model, it will be difficult for non-profits to outpace their corporate counterparts at \emph{training}---at least on their own \citep{Tan2025-dd}. Open-weight models, then, are a good start. While models like Llama might not be open-source in the strictest sense, researchers can download a snapshot of a particular released version with ``frozen" weights. These weights can be shared with other researchers. This specifically impacts the \emph{replicability} and \emph{reliability} of using LLMs in social science workflows. There are no guarantees that a specific proprietary model will continue to be accessible, and even more troubling, researchers find that OpenAI's models may even change within the same version \citep{Ollion2024-aa}. This stands to reason since, organizations like OpenAI are not guided by scientific pursuits: \emph{Our interests are not aligned}.\footnote{To be sure, this observation could be made about most proprietary software.}

\section{MODEL FOOTPRINT}

An important criticism of the contemporary LLM ecosystem is that it has only excerbated global energy consumption. The primary culprits are data centers, which amass data for training, train the models, and are used for cloud-based inference. Data centers accounted for approximately 2\% of energy use in the US in 2018, increased to 4\% by 2024, and is projected to increase to between 6.7\% and 12\% by 2028.  As data centers require electricity at all times and tend to be built in locations with limited access to wind or solar, they also use a larger share of ``dirty" energy than the typical household \citep{O-Donnell2024-st}. While data centers are also used for activities like streaming or messaging, GenAI is cited as key to rising energy costs \citep{Saul2025-fx} and an increase in water consumed when cooling data centers. However, it is difficult to know precisely the cost of GenAI in isolation, or even data centers in total, as estimating is both a technical and political challenge. Actors hide key details about this infrastructure \citep{Kainz2025-xp}, and a significant amount of hardware variation exists across data centers, including the energy efficiencies of the local power-grid.

One report estimated that one prompt to ChatGPT consumed ten times the energy of a traditional search engine search \citep{Aljbour2024-sd}, while internal Google researchers found recent improvements push that figure much lower for Gemini (\cite{Elsworth2025-dq}; see also \cite{Samsi2023-wj}). We are encouraged by efforts by, for example, Hugging Face's AI Energy Score\footnote{\url{https://huggingface.co/AIEnergyScore}} to pressure LLM providers to share details about energy consumption. We are also encouraged by recent efforts to focus on ``downscaling" and improving the performance of smaller language models \citep{Goel2025-cf, Wang2025-ce, Allen-Zhu2024-xs}. After all, it is not universally the case that larger (or even so-called state-of-the-art) models will outperform smaller or older models on all tasks \citep[e.g.,][]{Yu2023-np, Ollion2023-sh, Taylor2024-ee, Than2025-rq, Wright2025-ed}, and even computationally minimalist approaches like dictionary matching or reguluar expressions may be superior to LLMs for some tasks \citep{Stuhler2025-rp}.

In what follows, we introduce key considerations around model size and how this relates to computing resources when using LLMs for inferance. Our recommendation is to prioritize using the smallest viable model whenever possible.

\subsection{Parameters} 

In February 2019, OpenAI released a 124M parameter model, then a 355M model in May, a 774M model in August, and finally, the full 1.5B model in November of that year, dubbed GPT-2-Small, GPT-2-Medium, GPT-2-Large, and GPT-2-XL respectively. What do these numbers mean? Consider the 124M GPT-2 model. (See Appendix for a more comprehensive walk-through.) 

\begin{figure}
    \centering
    \includegraphics[width=\linewidth]{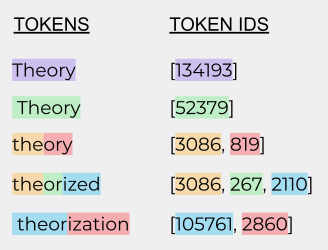}
    \caption{Example tokenization using OpenAI's tiktoken algorithm.}
    \label{fig:tokens}
\end{figure}

During tokenizing, we convert our prompt into token embeddings and position embeddings. These are initial states that summarize the meaning of these words in the context of the prompt. Here, tokens do not necessarily correspond to conventional ``words," but are rather common sequences of characters (see Figure \ref{fig:tokens}).\footnote{For instance, a word with a space in front of it may be considered a distinct token from the same word without that space, or with and without a leading capital. Furthermore, uncommon words (which also tend to be longer) are often tokenized into shorter subword sequences. Contemporary LLMs use a tokenization process based on byte-pair encoding or BPE \citep{Gage1994-tf}, but may even be tokenized as the underlying byte-level representation of each unicode character.}

The set of unique tokens forms a ``vocabulary," and GPT-2 uses a vocabulary of 50,257 tokens, which we designate $V$. Each token in our prompt is assigned a vector from the embeddings, and GPT-2 uses a vector that is 768 numbers long. 768 pops up a lot; we'll designate it $E$. The second embedding is used to assign a vector corresponding to the position of the token in the prompt, and GPT-2 uses a vector that is again $E$ numbers long and allows for 1,024 positions, or $P$. This means, we can only input a prompt of 1,024 tokens into GPT-2 (this is its \emph{context length}).

The token and position embeddings are passed to the Transformer blocks, and in GPT-2, there are 12 identical transformer blocks. Within each, there are a few sub-layers, but only one other parameter we'll need: the size of the ``hidden" layer. In the case of GPT-2, it is four times the size of the embeddings, or 3,072. We'll call it $H$. 

Thus, there are five components that define the total number of parameters: vocabulary size $V$, embedding dimensions $E$, context length $P$, number of transformer layers $L$, and the hidden layer's dimensions $H$. Along with the specifics of the model architecture (see Appendix), we can plug these into the following formula to calculate the total parameters $W$:


\begin{equation}
\begin{split}
W = & E(V + P) + L(12E^2 + 13E) + 2E
\end{split}
\end{equation}

\begin{equation}
\begin{split}
GPT2\ W =  & 768(50257 + 1024) + 12(12 *768^2 + 13 * 768) \\
           & + 768 * 2 \\
        = & 39,383,808 + 85,054,464 + 1536 \\
        = & 124,439,808
\end{split}
\end{equation}

We care about the number of parameters for two reasons: generally, (1) it has a strong relationship to the ``quality" of outputs (although, this depends) and (2) it has a direct relationship to the amount of computational resources required to use the model for inference.

Each parameter is a unique number that must be held in working memory (or RAM). This is a \emph{real} number, meaning we are dealing with decimals that could go on forever. But, computers cannot hold infinitely long numbers. Furthermore, all numbers must eventually be represented with two integers, 0 or 1 (in binary computers, which are the vast majority). So, we must use ``floating-point" numbers \citep{Goldberg1991-my}. This is a positive or negative integer multiplied by an integer power of some base that approximates the actual number. This results in a small amount of error between the actual number and its floating-point approximation. The base depends on how \emph{precise} our representations are, which with contemporary LLMs is usually 32-bit ``single precision" or less.\footnote{In modern ``binary" computers, a ``bit" is the most basic unit and is either a 1 or 0. Bits are ``chunked" into bytes, which is the smallest \emph{addressable} unit and is hardware-dependent. The common early computer architectures started with the 8-bit byte, and thus a byte conventionally refers to 8-bits. Contemporary machines are now based on 32- and 64-bit architectures. 32-bit representations are referred to as ``single precision," ``full precision," or ``float32," while 64-bit are referred to as ``double precision" or ``float64." Although 64-bit precision can represent more accurate numbers, most LLMs tend to be 32-bit or less.} This means that we only have 32 zeros and ones with which to represent the real number. 

To determine the bits of memory required by a model, we multiply the number of parameters by 32---often called ``bits per weight" or BPW. Thus, one way to reduce the working memory is to reduce the number of parameters. This is typically accomplished by reducing the size of the embeddings $E$, the number of transformer layers $L$, or the size of the hidden layer $H$. From the OpenAI report introducing the GPT-2 model, we can see that they vary $E$ and $L$ to train the different sized models \citep[]{Radford2019-bi}.\footnote{There are some discrepancies in various reports of parameter size: \url{https://github.com/openai/gpt-2/issues/209}. The largest discrepancy is for GPT2-Small, which was reported as having 117M parameters in the paper. Here, we use the parameter estimates from the developers page: \url{https://github.com/openai/gpt-2/blob/master/DEVELOPERS.md}} Their vocabulary is the same for all models, and each uses a $H$ that is four times their respective $E$.


Rather than just training a model with fewer parameters from scratch, smaller base models can be trained to mimic the performance of larger models through a compression process called \emph{distillation}, which uses a ``teacher" model to train a ``student" model. DistilBERT, for instance, is a distilled version of the Bidirectional Encoder Representations from Transformers (BERT) model \citep{Sanh2019-xc}. In general, the process looks a lot like fine-tuning (see the ``Model Architectures and Fine-Tuning" section later in this paper), where the student model is trained to produce outputs that look like the outputs of the teacher model. Along with pruning and quantization \citep{Kim2025-je}, this procedure is typically used internally to reduce the inference footprint of newly-trained large models. When DeepSeek-V3 was released, it made a huge splash as it was comparable in performance to GPT-4o, but purportedly cost about 5\% what OpenAI spent to train GPT-4o. OpenAI infamously accused of DeepSeek of ``inappropriately" copying outputs from GPT-4o to train the model---effectively distilling proper outputs from the model as training material \citep{Young2025-fi}.

\subsection{Quantization}

The default level of precision for most LLMs is 32-bits, meaning we have 32 ones or zeros to work with. For instance, 25.806 would be represented using three pieces: the \emph{sign} (positive or negative), the \emph{exponent}, and the \emph{mantissa}. Using the proper conversion, this number would look like the following:

\begin{center}
\begin{tabular}{r|ccc}
        & Sign & Exponent & Mantissa \\
size    &  1 bit  &   8 bits & 23 bits \\
binary  &  0   & 10000011 & 10011100111001010110000 \\ 
integer & +1   &    131   &     5141168 \\ \hline
        &      &          &             \\ 
value   & \multicolumn{3}{c}{25.805999755859375} \\
error   & \multicolumn{3}{c}{-0.000000244140625} \\
\end{tabular}
\end{center}

Using a less precise representation is another way to reduce the amount of working memory required. We could instead use 16-bits, for instance, all the way down to 1-bit \citep{Che2024-yt}. As we use fewer bits the amount of error between the actual and encoded \emph{value} will increase, and this \emph{might} lead to lower quality outputs (although not necessarily). This mapping of continuous value real numbers (like those in the weights of LLMs) to discrete representations is known as \emph{quantization}. If our model is quantized to 16-bits it will be much smaller (and faster) than at 32-bits. GPT-2 was originally at 32-bits precision, but we can calculate lower levels of precision, converted to gigabytes (GBs).\footnote{A standard ``byte" is 8 bits, and a ``gigabyte" is one billion bytes. For reference, as of 2025, popular smartphones tend to have somewhere between 4 GBs and 8 GBs of RAM and consumer laptops tend to have between 16 and 32 GBs of RAM, while desktops tend to have between 32 and 64 GBs. Universities with HPCs will commonly have machines with between 128 and 512 GBs of RAM, and a few may have up to 1024 GBs---a terabyte---of RAM. Beyond that, we run into supercomputer territory.}

\begin{equation}
\begin{split}
RAM &= 124\ M * 32\ bits = 3.968\ B\ bits = 0.496\ GBs\\
    &= 124\ M * 16\ bits = 1.984\ B\ bits = 0.248\ GBs\\
    &= 124\ M * 8\ bits  = 0.992\ B\ bits = 0.124\ GBs\\
\end{split}
\end{equation}

The main takeaway is that the lower the precision, the smaller the model and the faster it will run.\footnote{The above assumes every parameter is uniformly quantized at the same level. While there are now several different quantization procedures, far too many to summarize here, uniform quantization is the most common. Reducing the precision typically occurs after initial training (post-training quantization or PSQ), and may entail another round of training (quantization aware training or QAT) \citep{Gholami2021-ac}. Mixed-precision models, which quantize different parameters at different levels, have been shown to more effectively capture the ``signal" from the ``noise."} While it is typical to see outputs that appear ``lower quality," this quality difference is not directly related to the number of parameters \citep{Kumar2024-jr, Dettmers2022-vh}. Ongoing research in this space demonstrates that alternative quantizing methods can still lead to nearly identical outputs \citep{Frantar2022-ic}.

\subsection{Context Length}

The maximum number of tokens a model can process in one request is its \emph{context length} (or context window). This is the model’s attention span: it determines how much information the model can consider at once. Models like GPT‑3 can process about 2,048 tokens, while GPT‑4o, LLama-3.1 and Gemini-1.5 handle between 60,000 and one million tokens. 

While longer context lengths support tasks like summarizing long documents or maintaining multi‑turn conversations, they require more memory. Specifically, in standard architectures, if the context length is doubled, the amount of resource required for inference on the model will increase by four. An alteration of the position embeddings---called \emph{rotational} position embeddings or RoPE \citep{Su2021-id}---allows models to more efficiently expand the contexts, even up to millions of tokens. Even still, longer context is not always a boon \citep{Than2025-rq, Chae2023-yb}. 

As the context grows very long, various details might be ``forgotten" or have a weaker influence on the model's next output. The model's focus often skews toward more recent tokens (a kind of recency bias), which can cause it to overlook instructions or facts provided much earlier. Over long prompts, the middle portions can become diluted \citep{Liu2024-nb}. This phenomenon is sometimes called ``lost in the middle," ``context rot,"\footnote{\url{https://simonwillison.net/2025/Jun/18/context-rot/}} or context dilution.\footnote{This may be more of a feature of decoder-only architectures like GPT \citep[]{Fu2023-wx}.} In interactive settings, if a chat with an LLM goes on for many turns, the accumulated context can start to clutter the model's short-term memory. Even if nothing is explicitly forgotten, the sheer volume of text makes it harder for the model to maintain a clear, coherent thread. One partial solution is to first summarize or reduce longer texts to core ideas \citep[e.g.,][]{Waight2025-wr}.

\subsection{Computing Resources}

We argue that, to the extent social scientists use LLMs for research tasks, they should use the most computationally-efficient model (specifically, fewer parameters and/or quantized to lower precision) that performs well on manually-crafted benchmarks specific to their task.

There are two reasons we argue for selecting smaller models, both related to computational resources: access and energy. We presume that social scientists might have access to a range of computational infrastructures, from personal laptops and desktops to HPCs and cloud-computing services. If we use a smaller model for a given analytical task, more devices will be able to replicate that task. Furthermore, if we use a smaller model (fewer parameters or quantized to lower precision) it will require less energy to replicate that task as well---all things being equal \citep{Niu2025-yh, Alvarez2025-wu}.

Although researchers have discovered ways to make LLMs more efficient, due to the presumed relationship between scale and quality, some argue that there will be pressure to use efficiency gains to train and deploy larger models and thus not reduce overall energy use. However, as the initial cost involved in training new models is high, it is argued that incentives will lead to the widespread use of existing pretrained models and thus energy used in inference will likely outstrip energy used in training \citep{Niu2025-yh}.

From a researcher's perspective, inference is expended either \emph{locally} on a laptop or desktop, or \emph{remotely} through a university's HPC or industry-owned cloud-computing server---and in some cases, servers operated by national laboratories. ``Remote" means that we send our prompt from our own machine to another machine over an internet connection and the receiving machine expends the computational resources to produce the output, then sends the output back to our machine.\footnote{This is the case whether we use the ChatGPT web browser application, the ChatGPT smartphone app, or access GPT-4o using an API through our command-line interface.} Roughly, energy usage scales linearly with the number of tokens output \citep{Wang2025-yu, Alvarez2025-wu}.

We estimate that for local inference on consumer-grade GPUs, we could fit a full-precision model with between 2 and 3 billion parameters, or a model quantized to 4-bits with between 16 and 24 billion parameters.\footnote{For our purposes, we care about the CPU and its associated working memory (called RAM) and the GPU and its associated working memory (called VRAM). When using an LLM, we must load the entire model into working memory, either VRAM or RAM, or distributed across the two. We also know that the kinds of computations that LLMs perform (i.e., lots of small ones) works better on a GPU. Therefore, if possible, we want to use a GPU and load as much of the model into VRAM as possible. The most common GPUs that consumers have tend to be around 8-12GBs VRAM (according to data from the gaming platform Steam). These GPUs also tend to cost around US\$ 300.} What happens if the model size exceeds the memory available on a single GPU? We can split the model (called sharding), either between our CPU and the GPU, or across GPUs (if we have multiple GPUs). This splitting, however, comes with increased energy costs. Thus, we should select the smallest model that adequately performs our research task.

\section{TRAINING DATA}

\subsection{Size, Type, and Copyright}

LLM training data is \emph{big}---typically between a 200 billion and 20 trillion tokens. What types of texts show up in LLM training corpora?

A foundational data infrastructure for training most LLMs is the Common Crawl---a massive collection of over 300 billion webpages that grows by between 3 to 5 billion sampled pages every month, and maintained by a 501(c)(3) non-profit organization of the same name.\footnote{\url{https://commoncrawl.org/}} The Common Crawl consists of several petabytes of data, likely comprising 100+ trillion tokens. Webpages in the crawl are as diverse as the web: some of the most frequent web domains (as of December 2025) include blogspot.com, wikipedia.org, wordpress.org, ebay.com, europa.eu, fandom.com, and google.com.\footnote{\url{https://commoncrawl.github.io/cc-crawl-statistics/plots/domains}} Notably, the Common Crawl does not include webpages that require logins (thus a lot of social media is omitted). Languages range from English and Russian to Wolof and Zhuang, though the crawl is heavily dominated by languages with large linguistic communities---especially English, which usually accounts for nearly half of any month's crawl. The text is also disproportionately produced by younger men \citep{Luccioni2021-rf, Kuntz2023-aw}. Generally, a filtered version of the Common Crawl is used as it is filled with duplicated pages, non-text/HTML tags, and -- more importantly -- ``undesireable" content \citep{Luccioni2021-rf, Baack2024-vp}, such as forums filled with hate-speech and adult content webpages. Still, some of this content often makes it through to the training set \citep{Birhane2023-xs}.

But the Common Crawl is not all. Most groups, companies, and institutes training up LLMs diversify their training sets with other sources. Some of these training sets are proprietary; others are open source (see the ``Model Openness" section above). Open-source datasets such as The Pile added about 800 gigabytes of curated texts conducive to LLM use cases, such as medical publications, code repositories, and book collections. There are now many other large open-source corpora for LLM training, such as Zyphra's Zyda \citep{tokpanov2024zyda13tdatasetopen} and Ai2's Dolma \citep{soldaini2024dolmaopencorpustrillion}.

Even with open-source public training corpora, the question of copyright protection remains.\footnote{Copyright issues emerge even with the Common Crawl. The crawler doesn't log in to websites but may still be able to access content behind paywalls. But, the organization states that it respects requests to remove content. For instance, in 2023 \textit{The New York Times} requested that 16 million articles be removed -- and it seemed as though they complied. However, an investigator for \textit{The Atlantic} stated that many \textit{New York Times} articles appear to still be present \citep{Reisner2025-ug}} The Pile, for example, came under criticism for its inclusion of copyrighted works---most notably Books3, a collection torrented books that included both in-copyright and out-of-copyright works \citep{Tarkowski2024-qs, Tarkowski_2025}. The Pile was subsequently taken offline and replaced with a newer training dataset called The Common Pile---8 terabytes of text that is all public domain or with an open license---along with two new open-source LLMs trained on the data called the Comma models \citep{kandpal2025common}. Another training set purportedly free of copyrighted works is the Common Corpus \citep{langlais2025pleias}---a 2-trillion token multilingual collection free of any proprietary data (though see Kandpal et al. (\citeyear[p.~4]{kandpal2025common}) for a discussion of how The Common Pile's curation strategies might protect against license misatribution in a way that the Common Corpus does not). Finally, although there is ethical debate on using copyrighted materials for training LLMs, a U.S. District Court ruled \citep{bartz2025} that Anthropic using copyrighted works to train LLMs was fair use. However, the Court also ruled that Anthropic using pirated books was not justified by fair use; had Anthropic purchased access to the books, there would have been no issue. Furthermore, some argue \citep{Doctorow2025-zq, Tarkowski2024-qs} that increasing barriers to accessing content for training is unlikely to benefit creative workers, and much more likely to benefit large corporations.

\subsection{Human and Synthetic Data}

Collecting data for pre-training (and even fine-tuning, as we discuss later) is computationally, socially, and fiscally expensive---so much so that researchers have started exploring using synthetic data for such tasks. \emph{Synthetic data} in this sense is data that is itself generated from an LLM and is then used as input to further train that (or another) LLM \citep{zhang2025does}. Reseachers have explored using synthetic data in ``data-scarce" environments, such as creating LLMs that are better representative of languages with relatively less prevalence in naturally-occuring training corpora such as Greek \citep{li2024culturellm}.\footnote{Using synthetic training data is distinct from the related practice of \emph{in-silico sampling}, where an LLM is used to simulate survey responses from real-world seed survey data \citep[see, e.g.,][]{kozlowski2024silico}. This use case has its own issues, however \citep{boelaert2025machine}.}

Using synthetic data for training has some major pitfalls, however---at least for social science applications where performance on benchmarks is not the ultimate litmus test. Most notable is the tendency for training with synthetic data to create a sort-of regression to the mean where generated texts converge on modal patterns and minimize variance we would expect to see in natural training data. Synthetic data patterns tend to be more uniformly distributed, leading to overfitting and poorer instruction-following capabilities for models trained on them \citep{chen2024unveiling}. Iterative training with synthetic data---where later generation models are trained on data produced by earlier generation models that were themselves trained on synthetic data---compounds this problem, leading to successively lower-variance pattern distributions that increasingly fail to mimic real-world data \citep{seddik2024bad}. This effect, known as \emph{model collapse}, leads to poor model performance on downstream tasks \citep{seddik2024bad,shumailov2023curse}. There is some evidence that certain ratios of real-world and synthetic data can mitigate this problem and increase computational efficiency \citep{kang2025demystifying}. Yet, training solely on synthetic text seems to, at best, propogate (as opposed to magnify) errors and biases \citep{zhang2025does} and erase linguistic diversity at worst \citep[p.~4]{seddik2024bad}.

\subsection{Fine-Tuning Datasets}

Fine-tuning involves adapting a foundational LLM---itself pre-trained on large, general training corpora of the variety discussed above---to better position the model either for a more generic stylistic preference or for a specific task (discussed in more detail in the next section). Chae and Davidson (\citeyear{Chae2023-yb}), for example, fine-tuned a series of general LLMs for political stance detection. The goal was to isolate the politicians referenced in social media messages and the posters' stances held towards them (e.g., in favor, against, or neutral). The authors did this by providing the models with hand-annotated datasets of select social media posts before passing their ``held-out" data---the posts not used in the fine-tuning---to the fine-tuned models.

Fine-tuning datasets typically need to be large---indeed, Chae and Davidson (\citeyear{Chae2023-yb}) found that fine-tuned models tended to perform better the more fine-tuning examples provided \citep[see also][]{Zhang2024-en}. These datasets are basically identical to benchmark datasets in the sense that they require annotations of example texts---usually (but not always) \emph{human} provided---indicating a preferred output. This introduces not just questions of fiscal cost, but social and ethical dilemmas as well. Annotation dataset construction often relies on hidden, underpaid, and underrecognized labor \citep{paullada2021data, Bender2025-wa}. Recent work has systematized how the machine learning community can better facilitate dataset development in socially responsible ways that incorporate and reward local data workers as research collaborators \citep[e.g.,][]{paullada2021data}, but researchers should be ever-vigilant about the dataset construction processes undergirding any ``off-the-shelf" fine-tuned LLMs.

\section{MODEL ARCHITECTURES AND FINE-TUNING}

Currently, there are three main types of LLMs---encoder-only, decoder-only, and encoder-decoders---which are all based on the Transformer architecture. After discussing these three types, we describe how intial models may be further ``fine-tuned" for specific tasks, and finally how so-called ``reasoning models" and retrieval-augmented generation models (RAGs) automate the curation of the generation process.

\subsection{Encoders, Decoders, and Encoder-Decoder Models}

The original Transformer architecture included two main components or ``stacks," one labeled the ``encoder" and the other labeled the ``decoder" \citep{vaswani2017attention}.  Subsequent work building on the Transformer discovered that using each in isolation can be useful. For instance, both BERT and GPT are language models based on the Transformer, but BERT only uses the encoder stack of the Transformer while GPT only uses the decoder. This etymology can be confusing, as encoder could mean many related things in the context of machine learning---but following this literature, it specifically refers to the encoder and decoder portions of a Transformer. This discussion can also be confusing as the encoder and the decoder stacks are \emph{nearly} identical.

The encoder stack lets each word attend to every other word in a sequence, capturing contextual relationships across the entire input. BERT is a well-known encoder-only system. It is pretrained to predict masked words (i.e., it can see all of the words around a target word it is trying to predict) and next sentence prediction. When inputing text into BERT, the  output is not the next word in a sequence---i.e., it is not technically ``generative." The output is a set of embeddings which represent the meaning of each token in the input sequence. An analyst can use these embeddings for various downstream tasks, including generating a next word, but this model tends to excel for classification tasks.

Decoder-only systems do not ``learn" anything new from the tokens in the input \emph{per se}. Rather, a decoder-only model maps the words onto its internal representations and uses that to predict the next token before feeding that token along with the input back into the model and predicting the next token, and so on---a property known as autoregression. Although nearly identical to a encoder, decoders include one different layer within the stack: masked self-attention, which prevents them from ``peeking" ahead to see the words that might come after a predicted token (sometimes called \emph{causal masking}). Thus, unlike BERT which is bidirectional, GPT is unidirectional and can only use past tokens to predict a next token. The decoder-only also appends a final layer that outputs a single token.

Finally, some systems combine both components. In these \emph{encoder-decoder} models, the encoder reads the input text, creates an embedding represenation of each token, passes that representation to the decoder, which uses it to generate an output token conditioned on the encoder’s representation. This approach is particularly common for tasks like machine translation and summarization, where the output depends on the entire input. Popular encoder-decoder models include T5, BART, and MASS. During pre‑training, these models learn to reconstruct corrupted or masked spans of text.

For example, the original Transformer uses an encoder to read the source sentence (say, in French) and a decoder to produce the target sentence (say, in English). The encoder can attend to all parts of the input simultaneously (no causal masking), creating a rich, contextual embedding of the input. The decoder then uses this embedding (via cross-attention) along with what it has generated so far to produce the final output. This two-stage design explicitly separates \emph{understanding} the entire input from \emph{generating} the output. Encoder-decoder models are often used when we want to transform input text into a different output (translation, summarization), where having a full encoding of the input can be advantageous.

\subsubsection{Prompting: Few-Shot to Zero-Shot}

In a generative LLM (i.e., decoder-only or encoder-decoder models), the user inputs a chunk of text called a \emph{prompt}. This prompt tells the model where to begin generating new text. The ultimate goal for developers in this space is to create a model that can always return precisely what the user wants (whatever that may be). This is called \emph{zero-shot}. For instance, if our prompt reads: ``Is the following sentence positive, negative, or neutral? Sentence: I hate mondays!" We want the model to output ``negative," without any further tweaking of the model. In practice, this works some of the time. For some tasks, we may provide a few examples within our prompt, called \emph{few-shot}. For example, if you want the model to answer trivia questions, your prompt might include: 
\begin{quote}
Q: What is the capital of France? A: Paris
Q: Who wrote \emph{Romeo and Juliet}? A: William Shakespeare
Q: [new question]…
\end{quote}
Few-shot examples may also include formating or possible steps to solve a problem. There are now several techniques for organizing prompts in ways to better corral LLMs to produce desired outputs. For instance, we might include text at the beginning of our prompt indicating that the LLM is a ``helpful research assistant" in the hope that it will succinctly answer our requests (we outline several strategies in the Appendix).

\subsection{Fine-Tuning: Instruct, Preference, and Reasoning}

\begin{table*}[]
\centering
\begin{tabular}{p{3cm}|c|c|c}
\toprule[1.5pt]
               &     Model            &  Base     &  Fine-Tuning      \\
Provider       &     Name             &  Model    &  Task            \\ \hline
Databricks     &     Dolly            &  Pythia      &  Instruct          \\
Apple          &     OpenELM-Instruct &  OpenELM     &  Instruct          \\
Stanford       &     Alpaca           &  Llama       &  Instruct          \\
Ai2            &     Tülu             &  Llama       &  Instruct          \\
DeepSeek       &     DeepSeek-R1      &  DeepSeek-V3 &  Reasoning          \\
OpenAI         &     o1               &  GPT         &  Reasoning        \\
Alibaba        &     QwQ-32B-preview  &  Qwen        &  Reasoning        \\
IBM            &     Granite 3.2      &  Granite     &  Instruct/Reasoning  \\
\bottomrule[1.5pt]
\end{tabular}
\begin{center}
\caption{Selection of Fine-Tuned Model Families}\label{tab:terms2} 
\end{center}
\end{table*}

Pretrained LLMs are initially trained on vast amounts of general text. To adapt these models for a specific domain or task, or to get them to respond in perferred styles, we can perform \emph{fine‑tuning}. This process continues the training process, but now on a smaller, \emph{labeled dataset}. Fine‑tuning allows the model to adjust its internal parameters, while still using the general knowledge gained during pre‑training. While we could use fine-tuning far any number of tasks \citep{Chae2023-yb}, it is common to find off-the-shelf LLMs that have been tuned to \emph{follow instructions}, \emph{align with human preferences}, and \emph{engage in ``reasoning"} (see Table \ref{tab:terms2} for some example fine-tuned models).\footnote{Following the procedures similar to instruction- or preference-tuning, an analyst could further alter the parameter weights of the model for more specific tasks, just as Chae and Davidson (\citeyear{Chae2023-yb}) do for political stance detection. On larger LLMs, this typically involves an efficient training process like QLoRA \citep{Dettmers2023-lr, Han2024-eq, Zhang2024-en} to update the model weights. When fine-tuned on highly specialized tasks, however, these models predictably lose flexibility. Prefix-tuning \citep{Li2021-kl} and prompt-tuning \citep{Lester2021-vp, Li2025-wb}, and variations on these \citep{Liu2022-df, Han2024-eq}, are procedures that use a similar fine-tuning process, but instead of updating all the model weights, only update special vectors which are concatenated to the model. With prefix-tuning, the vectors are concatenated to every block. With prompt-tuning, the vectors are only concatenated to the input embeddings.}

\emph{Instruction tuning} is a more recent training method that uses fine‑tuning to make the model better at a very general task: following diverse instructions \citep{Ouyang2022-if}. Instead of providing only input–output examples, the training data includes a written description of the task itself (the ``instruction").  For example, a training example may consist of an instruction such as ``Translate the following sentence into French,'' the input sentence, and the desired translation. By learning from such triplets, the model becomes more adept at understanding what a user wants and can generalize to new instruction-following tasks not seen during training. While these instructions could be included in every prompt, fine-tuning both allows the model to save tokens on generation (see the ``Context Length" section above) and also demonstrates outputs more aligned with the task of instruction-following than if the instructions were only included in the prompt. Instruction‑tuned models, like OpenAI's InstructGPT and Google’s FLAN family, seem to require fewer examples to improve performance on various requests \citep{Wei2021-jj}.

\emph{Preference tuning} is a type of fine-tuning that ``rewards" models when it produces an output that aligns with human \emph{preferences}---i.e., reinforcement learning with human feedback, or RLHF \citep{Ziegler2019-eo, Askell2021-kp, Bai2022-kf}. To build preference training data, users are presented with two possible responses from a given prompt and asked to choose which they prefer. For example, they may be asked which response is more ``helpful" or asked which is more ``harmless" (tasks often falling under the heading ``alignment"). With enough pairwise comparisons, multiple outputs can be ranked from most to least helpful (or harmless) for a given prompt. This rank provides a ``reward," which the model wants to maximize over the fine-tuning process, aligning it with ``human" preferences. This, of course, raises important questions \citep{Martin2023-mk, Zajko2025-pw}, in particular: whose preferences? 

\emph{Reasoning-tuned} models are another common flavor of off-the-shelf fine-tuned LLMs. A variation on few-shot prompting is Chain-of-Thought (COT) prompting \citep[]{Wei2022-is}, where instead of simply providing the correct answer to an example, we also provide the steps one may take to arrive at that correct answer (see Appendix). In fact, even adding ``Let's think step by step" to a prompt can improve model outputs \citep{Kojima2022-le}. Building on these insights, reasoning models are LLMs that have been further fine-tuned using reinforcement learning to generate this \emph{chain-of-thought} styled output automatically before arriving at a final output. In other words, just as instruction-tuning is a method to ``absorb" instructions that might be included in a prompt into the model weights, reasoning models absorb COT-style prompting. While using a reward similar to preferencet-tuning, tweaking an LLM to reason can often involve a task with a verifiable output, such as solving a math question or completing a code function, and may even involve verifying each step in the reasoning process \citep{Lightman2023-jv}. The DeepSeek-R1-Zero model, for example, was fine-tuned to put the ``reasoning" output inside special \texttt{<think>...</think>} tags and the final output in between \texttt{<answer>...</answer>}, and penalized when it did not do so. Thus, reasoning models are simply auto-generating a prompt that further \emph{constrains} the generation process. While this may improve performance on some tasks, it also necessarily comes with increased computational costs.\footnote{We should be careful not to anthropomorphize the model and assume that the output in the ``reasoning" portion of the output---its ``reasoning trace"---is actually how the model ``thinks" or are steps the model actually took to arrive at a conclusion \citep{Chen2025-yr}. }

\subsection{Retrieval-Augmented Generation}

Most LLMs have incorporated Wikipedia during the pretrain process and therefore include much information from various articles in the model weights. We could, therefore, ask a factual question, such as how old an actor would be today. But, what if the actor recently passed away? The relevant Wikipedia entry the model trained on would be out of date. We could, then, navigate to that Wikipedia page and copy/paste the text from that full page into our prompt, along side our request. The LLM could use this updated information to generate its response, and would be more likely to be accurate. But, what if, instead of the user retrieving the relevant page, the model was able to automatically search and retrieve this page and incorporate it into our prompt? This is precisely what Retrieval-Augmented Generation models (RAGs) or Retrieval-Augmented Language Models (RALMs) accomplish \citep{Ram2023-wj, Gao2023-uu}.

When prompted, the model first engages in a retrieval step to grab existing documents related to the prompt. This could be done by searching a database, using a search engine, or querying a vector index of text embeddings to find passages that semantically match the query.\footnote{Retrieval has been well-studied for decades and traditional techniques like the Okapi BM25 \citep{Sparck-Jones2000-sz, Robertson2009-uo} developed in the 1970s-80s seems to be well-suited for this task \citep{Ram2023-wj}. However, vector-based retrieval systems are also common, which use an encoder model to create embedding representations of both texts and queries.} The top relevant pieces of text are then fed into the prompt alongside the original question. Now the model has, in its context window, some real background information to draw from. Furthermore, Wright et al. (\citeyear{Wright2025-ed} find that RAGs can somewhat mitigate the tendency for LLMs to homogenize \citep{Zhang2025-st}. As the retrieved text is prepended to the prompt, however, RAG models run into the problem of context length and dillution (as discussed previously). A strength of RAG models, though, is that they allow the source of outputs to be traced back to a verifiable source. These models are thus promising for scholars who work with collections of text---such as interview transcripts, fieldnotes, or archival materials---which could be queried from a relatively small, locally-operated model.

\section{CONCLUSION}

As we stated at the onset, when deciding to use a language model, there are numerous models to choose from. How do we select among language models? We explored four categories of consideration when selecting language models: (1) model openness, (2) model footprint, (3) training data, and (4) model architecture and fine-tuning. We argue that should social scientists have a task well-suited for an LLM, they should begin with smaller models, prioritize using the most open models available, trained on text that is public domain or open-license, and should create validations for the specific task for which they are using the model. Among other things, smaller models can be used for inference on a wider variety of hardware, making reproduciblity more accessible. Open-weight models allow researchers greater control over data storage---potentially a requirement when studying sensitive data. And, fully-open models allow us to better understand the relationship between training data, training procedures, and model outputs (increasing reliability and replicability), while also making it difficult for developers to rely on hidden labor.

Different model architectures, fine-tuned in various ways, also afford different analytical possibilities. Consider two scenarios involving transcripts of political speeches: (1) a researcher wants short summaries for each transcript to help organize their initial analysis; and (2) another researcher who wants to know which transcripts engage with a range of possible themes along with possible direct quotes. Both tasks would suggest a model with decoding capabilities, and one fine-tuned to follow instructions. For the first task, there is a strong need to associate a summary with a specific interview; thus the researcher would feed the exact transcript along with the prompt to summarize. They would need to be sure that the transcripts do not outstrip the context length (and may even benefit from pre-chunking the transcripts). For the second task, the researcher would benefit from a RAG model that can first retrieve relevant transcripts based on a request, and then provide quotes from a specific transcript. Both provide ways of swiftly surveying the entire corpus of transcripts; however, the researcher can ground the model outputs in the actual transcripts and easily construct benchmarks to validate the outputs. If need be, they could further fine-tune the model on high quality input-output pairs constructed by the researcher.

Although much has been said about the \emph{scale} of tasks---in terms of speed or time---enabled by large pretrained language models, we are ambivalent. Simply doing more, faster is not clearly a boon for sociology or academia. As stated at the onset, the preceding discussion is applicable to the class of tasks that analysts would like to iterate and where such iteration directly relates to scientific explanation. We argue that it is \emph{iteration} where language models can supplement careful, critical, scholarly analysis. Indeed, we argue that iteration is a key strength of all computational methods: we can perform a task, and, with high-fidelity, reproduce that same task. If we share our materials, other researchers could also reproduce this task. Iteration also affords observing the consequences of minor changes to the workflow, small experiments, and slight changes in perspective. Proprietary large language models---with opaque training procedures, concealed training data, undocumented changes to models---are a hinderance to iteration.

\bibliographystyle{SageH}
\bibliography{llms_refs.bib}

\appendix

\section{APPENDIX}

\subsection{GPT-2 Model Architecture Walk-through}\label{app:gpt}

To better understand the basic size of Transformer-based models, specifically decoder-only models, let's walk through the GPT-2 model.

As discussed in the main text, during the process of tokenizing, we convert our prompt into a token embedding and a position embedding. The unique tokens are referred to as a "vocabulary," and GPT-2 uses a fixed vocabulary of 50,257 tokens, which we designate $V$. Each token in our prompt is assigned a vector from the embeddings, and GPT-2 uses a vector that is 768 numbers long. A separate embedding is used to assign a vector corresponding to the position of the token in the prompt, and GPT-2 uses a vector that is again 768 numbers long and allows for 1,024 positions. This means, we can only input a prompt of 1,024 tokens into GPT-2. 768 is going to pop up a lot, we'll designate it $E$. 

The token and position embeddings are passed to the transformer blocks, and in GPT-2, there are 12 identical transformer blocks. Within these there is one other parameter we'll need which defines the size of the "hidden" layer. In the case of GPT-2 it is four times the size of the embeddings or 

These provide the first set of parameters: 

\begin{equation}
Tokens = V * E = 50257∗768 = 38,603,776
\end{equation}

\begin{equation}
Positions = Positions * E = 1024∗768 = 786,432
\end{equation}

Next, we have the transformer blocks. In GPT-2 there are 12 transformer blocks which are identical. The first layer in a block is a "norming" layer which produces two vectors equal to the size of the token and position embeddings, which for GPT-2 is 768:

\begin{equation}
Norm_1 = 2 * E = 2∗768 = 1536
\end{equation}

The next layer in a block is the "attention" layer. This layer is a bit more involved, with four sub-layers. The first sub-layer is a matrix that is 768 by 2,304 dimensions representing three pieces of information that were input to the attention layer (known as the query, key, and value). The three input vectors are the same size as the embeddings, which for GPT-2 is, again, 768 (which by 3 is 2,304). Finally, this layer includes three bias vectors for each input. 

\begin{equation}
\begin{split}
Attention_1 &= (E * (3 * E)) + (3 * E) \\
            &= 768 * 2,304 + 2,304 = 1,771,776
\end{split}
\end{equation}

The second sublayer is involved in merging information and produces a matrix of 768 by 768. It also includes a bias vector that is also 768 dimensions.

\begin{equation}
\begin{split}
Attention_2 &= E * E + E = 768 * 768 + 768 \\
            &= 590,592
\end{split}
\end{equation}

There are two more sublayers in GPT-2's attention layer, but these are not included in the parameter calculation. Therefore, the total parameters for the attention layer involve adding the parameters from the first two sublayers.

\begin{equation}
\begin{split}
Attention = &Attention_1 + Attention_2\\
          = &1,771,776 + 590,592\\
          = &2,362,368\\
\end{split}
\end{equation}

Next up, we have another norm layer, which is identical to the first:

\begin{equation}
Norm_2 = 2 * E = 2∗768 = 1536
\end{equation}

Finally, we have the ``feed forward" layer that includes a few sublayers. The first layer involves a "hidden" matrix that is the length of our embedding vectors by $H$ dimensions long. $H$ just happens to be four times the length of our embeddings. We will also have a bias vector that is $H$ numbers long.

\begin{equation}
\begin{split}
Feed\ Forward_1 &= E * H + H \\
                &= 768 * 3072 + 3072 \\
                &= 2,362,368
\end{split}
\end{equation}

The second sublayer projects the prior layer's output into a matrix of size $H$ by $E$, and also includes a bias vector that is $E$ numbers long.

\begin{equation}
\begin{split}
Feed\ Forward_2 &= H * E + E \\
                &= 3072 * 768 + 768 \\
                &= 2,360,064
\end{split}
\end{equation}

Again, there are two more sublayers in GPT-2's feed-forward layer, but these are not included in the parameter calculation.

\begin{equation}
\begin{split}
Feed\ Forward &= Feed\ Forward_1 + Feed\ Forward_2 \\
              &= 2,362,368 + 2,360,064 \\
              &= 4,722,432
\end{split}
\end{equation}

Then, we have one final norming layer:

\begin{equation}
Norm_3 = 2 * E = 2∗768 = 1536
\end{equation}

We would then multiply the total transformer block by 12 for each block in GPT-2. From this walk through, we can see there are five main components that define the total number of parameters $W$: vocabulary $V$, embedding dimensions $E$, sequence lengths $P$, number of transformer layers $L$, and the hidden layer dimensions $H$. This produces:
\begin{equation}
\begin{split}
W = & (V*E + P*E) + L(2E + (4E^2 + 4E) + 2E \\
    &  + (2E * H + E + H)) + (2E) \\
  = & E(V + P) + L(12E^2 + 13E) + 2E
\end{split}
\end{equation}


\begin{equation}
\begin{split}
GPT2\ W  = & 768(50257 + 1024) \\
            + & 12(12 *768^2 + 13 * 768) \\
            + & 768 * 2 \\
        &= 39,383,808 + 85,054,464 + 1536 \\
        &= 124,439,808
\end{split}
\end{equation}

\subsection{General Prompting Strategies}\label{app:prompt}

\emph{Zero-shot prompting}. Zero-shot entails a prompt that request the model produce a specific output, but without giving any examples. For instance, we might write: ``Summarize the following article:" and then provide the article text. The model has to figure out what we want (a summary) and produce it without demonstrations. While zero-shot prompting may appear to work well for a single request, we might find that it is less successful when repeating the task. For example, asking it to summarize several articles, one after the other. Even with zero-shot, however, we will often find we need to ``tweak" the phrasing of the request. 

\emph{Few-shot Prompting}. In few-shot prompting, we provide the model with examples of the task \emph{within the prompt itself}, before asking it to perform the task on a new query. For example, if we want the model to answer trivia questions, our prompt might include: 
\begin{quote}
Q: What is the capital of France? A: Paris
Q: Who wrote \emph{Romeo and Juliet}? A: William Shakespeare
Q: [new question]…
\end{quote}
Then the model will continue with the answer. Few-shot examples prime the model by showing it the format and possibly the steps to solve a problem. The model uses the patterns in the provided examples to infer how it should respond. The downside is that it makes the prompt longer (using up part of the context window).

\emph{Role Prompting (Persona or Context Setting)}. This strategy involves telling the model to adopt a certain role, tone, or point of view, which can guide the style and detail of the response. For instance, we might begin the prompt with something like: ``You are a friendly librarian who speaks in simple language..." or ``Act as a strict grammar teacher correcting a student's essay." By giving the model a role, we provide implicit instructions about how the answer should sound and what details to include. Role prompting can influence the output style: the model might use more technical language if we say ``You are a software documentation assistant," or add more explanatory detail if instructed to be a teacher. We're setting the scenario, which can help maintain consistency in tone and also avoid certain behaviors.

\emph{Formatting Instructions and Style Guidelines}. We often want tell the model *how to structure its output*. LLMs are capable of producing output in various formats (paragraphs, bullet points, tables, JSON, CVS, code snippets, etc.), but won't always guess the format we need unless we specify it (note that some models have been fine-tuned to produce structured outputs). For example, if we want an answer in a bulleted list, we might write, ``Give the answer as a bullet-point list of the main advantages." If we need a JSON object as output (for easy parsing), we can instruct, ``Respond with a JSON object containing these fields: ...". By guiding the format, we make it easier to extract information and ensure the answer meets our needs.

\subsection{Reasoning-Enhanced Prompting Strategies}

For complex problems (math, logic puzzles, multi-step reasoning), it helps if the model can work through the problem step by step. Researchers have introduced several prompting techniques to encourage this kind of ``reasoning."

\emph{Chain-of-Thought (CoT)}. This technique involves prompting the model to produce a step-by-step explanation or thought process before giving the final answer. For example, instead of asking directly ``What is the answer?", we might say ``Let’s think this through step by step:" and then have the model outline its reasoning. By explicitly generating intermediate reasoning steps, the model often arrives at more accurate answers for complex tasks, specifically math word problems and logic puzzles. CoT can be zero-shot (just instructing the model to think aloud) or few-shot (providing examples of how to reason and then solve).

\emph{Tree-of-Thought (ToT)}. While chain-of-thought is a single linear sequence of reasoning, Tree-of-Thought goes a step further by prompting the model to explore multiple possible solution paths. In a Tree-of-Thought approach, the model might generate several different possible steps or ideas at a given point (branching out like a tree), then evaluate or choose among them, and continue this process iteratively. The model isn't confined to one train of thought but can consider alternatives. Implementing ToT often requires an algorithmic wrapper around the LLM---for instance, prompting the model for possible next steps, scoring them, and searching through the ``tree" of possibilities. The result is a more deliberate problem-solving process that can improve results on tasks that require exploration (like puzzles or games).

\emph{Self-consistency Decoding}. Instead of asking a question once and taking the first answer, with self-consistency we prompt the model to generate multiple reasoning chains (multiple answers) by sampling different possible outputs (for example, by using a higher randomness/temperature in generation). Each of these runs will produce a chain-of-thought and a final answer. The idea is that while any single run might make a mistake in reasoning, by looking at a collection of answers we can pick the answer that appears most frequently or seems most consistent across the various runs. In practice, one might have the model answer the same question, say, 5 or 10 times. The correct answer, if the model is capable of finding it at all, will show up in the majority of those attempts (because there are many ways to be wrong but fewer ways to be correct). By choosing the answer with the highest agreement among the samples, we boost accuracy.

\end{document}